\documentclass{article}
\pdfoutput=1

     \usepackage[preprint]{arXiv}

\usepackage[utf8]{inputenc} 
\usepackage[T1]{fontenc}    
\usepackage{hyperref}       
\usepackage{url}            
\usepackage{booktabs}       
\usepackage{amsfonts}       
\usepackage{nicefrac}       
\usepackage{microtype}      

\usepackage{color}
\usepackage[ruled,vlined]{algorithm2e}
\usepackage{algorithmic}
\usepackage{caption}
\usepackage{graphicx}
\usepackage{amsthm,amsmath,amssymb,mathrsfs}
\usepackage{float}
\usepackage{color}
\usepackage{eucal}
\usepackage{amssymb,amsmath,amsthm}
\usepackage{dsfont}
\usepackage{multirow}
\usepackage{natbib}
\usepackage{subfig}
\usepackage{tikz}
\newcommand{\minipagett}[4] { 
\begin{minipage}[t]{#1\linewidth}\vspace*{0pt}#3\end{minipage}%
\hfill%
\begin{minipage}[t]{#2\linewidth}\vspace*{0pt}#4\end{minipage}}

\newcommand{\alg}{CLEIR-Net}

\title{Adaptive County Level COVID-19 Forecast Models: Analysis and Improvement}

\author{%
  Stewart W Doe \thanks{Authors contributed equally to this paper and are listed alphabetically} \\
  Department of Computer Engineering\\
  University of Maine\\
  Orono, ME 04469 \\
  \And
Tyler Russell Seekins $^*$ \\
  Department of Chemical Engineering\\
  University of Maine\\
  Orono, ME 04469 \\
  \And
 David Fitzpatrick \\
  Department of Engineering Physics/Mathematics\\
  University of Maine\\
  Orono, ME 04473 \\
  \And 
  Dawsin Blanchard \\
  Department of Computer Science\\
  University of Maine\\
  Orono, ME 04473 \\
  \And
  Salimeh Yasaei Sekeh \thanks{Corresponding author - Email: salimeh.yasaei@maine.edu} \\
  Department of Computer Science\\
  University of Maine\\
  Orono, ME 04473 \\
}

\begin{document}

\maketitle

\begin{abstract}

Accurately forecasting county level COVID-19 confirmed cases is crucial to optimizing medical resources. Forecasting emerging outbreaks pose a particular challenge because many existing forecasting techniques learn from historical seasons trends. Recurrent neural networks (RNNs) with LSTM-based cells are a
logical choice of model due to their ability to learn temporal dynamics. In this paper we adapt the state and county level influenza model, TDEFSI-LONLY, proposed in \cite{Wangetal2020} to national and county level COVID-19 data. We show that this model poorly forecasts the current pandemic. We analyze the two week ahead forecasting capabilities of the TDEFSI-LONLY model with combinations of regularization techniques. 
Effective training of the TDEFSI-LONLY model requires data augmentation, to overcome this challenge we utilize an SEIR model and present an inter-county mixing extension to this model to simulate sufficient training data.
Further, we propose an alternate forecast model, {\it County Level Epidemiological Inference Recurrent Network} (\alg{}) that trains an LSTM backbone on national confirmed cases to learn a low dimensional time pattern and utilizes a time distributed dense layer to learn individual county confirmed case changes each day for a two weeks forecast.  We show that the best, worst, and median state forecasts made using CLEIR-Net model are respectively New York, South Carolina, and Montana.

\end{abstract}

\section{Introduction}

The ongoing novel COVID-19 pandemic has strained United States healthcare systems increasing the importance of optimizing medical resource allocation. This optimization will depend on the accuracy and precision of the models used to forecast confirmed cases in a given area \cite{allocation, communication,optimisation,EEA_forecast, supply}. To the best of our knowledge, current COVID-19 modeling literature primarily consists of two main approaches: SEIR modeling and Recurrent Neural Network (RNN) forecasting national trends for the purpose of determining transmission parameters and evaluating the effectiveness of national policies \cite{transmission_hubei, R_hubei,Hubei, Brazil, Canada, mlworld}. The SEIR model has a long history in epidemiology \cite{Kermack_model} and the benefit of this model is the direct mathematical modeling of disease transmission which can be fit to known data 
as was seen at the outbreak in Wuhan \cite{transmission_hubei, R_hubei,Hubei,Anastassopoulou2020}. The SEIR model is particularly useful for policy makers as the measurable constants that affect the transmission and disease dynamics can be easily monitored. Its vulnerability is its lack of complexity to represent the full range of the pandemics transmission dynamics. This becomes evident when forecasting at the county level in an interconnected region like the United States.

Recurrent Neural Networks (RNNs) are known to perform well when learning sequential patterns and are often applied to capture dynamic temporal behavior in time series. Similar to other types of deep networks, RNN training requires a large number of data to handle the high complexity associated with dimensionality. Possibly the largest challenge in forecasting COVID-19 is overcoming the curse of dimensionality with the limited data available. Here, there is simply not enough data to justify a complex model with millions of parameters and explanatory features. One solution is using simulated data from a deterministic epidemiological model to supplement real data. For simplicity of the process we may simulate epidemics using randomized but relevant parameter sets. The results are then used to train an RNN model. 

We adapt a variant of an influenza epidemic forecast model, TDEFSI, proposed by \cite{Wangetal2020} to forecast the COVID-19 pandemic in county level. TDEFSI features a single branch RNN with two-stacked LSTM layers to capture the time pattern in the times series training data. The authors utilize the technique of data augmentation: the generation of SEIR simulated data to be used as training data.
However, the standard SEIR models neglect migration to and from the modeled region which might change the proportion of the labeled populations. This is not a suitable assumption when modeling individual counties since cross county flows remain significant even under lockdowns in urban areas such as New York City due to their interconnection and close proximity. In our paper, this is amended by incorporating an inflow and outflow term into each labeled population's Ordinary Differential Equation (ODE). 

In this work, we propose a novel, state-of-the-art, architecture of RNN, {\textit {County Level Epidemiological Inference Recurrent Network}} (\alg{}), that requires fewer parameters, thereby mitigating overfitting, and, considerably improving both forecast error and training time. The approach reduces the number of parameters using a hierarchy of relationships to learn a mapping from a low dimensional, national level, signal to a high dimensional, county level, signal. 
\vspace{-0.2cm}
\section{Related Work}
\vspace{-0.2cm}
From the onset of the pandemic epidemiologists began using deterministic SEIR modeling in Wuhan, China to estimate the basic reproductive number and forecast infections \cite{transmission_hubei,R_hubei,Hubei,Anastassopoulou2020}. Several teams have used machine learning models to produce national forecasts of confirmed cases in Canada, Brazil, Wuhan, Italy, South Korea, and the United States, as found in \cite{Brazil, Canada, mlworld}. During the writing of this paper we found \cite{SEIRNet} used a combined LSTM network and SEIR architecture with mobility data to model county level confirmed cases for SEIR parameter estimation and measuring the effect of population mobility. At the time that we started this project there were no published models used to forecast high resolution county-level data for COVID-19 to date. This led us to adapt forecasting models used for other epidemics, specifically the TDEFSI, described in \cite{Wangetal2020}.

Three variants of the TDEFSI models were developed to forecast influenza epidemics. Two of these utilize the two branch architecture and provide superior results. However, the models require previous seasons data to train making them unusable for the COVID-19 pandemic due to its novelty. The third variant is TDEFSI-LONLY with a one branch structure. This model along with proposed regularization terms were trained on SEIR simulated New Jersey data and outperformed 
LSTM[\cite{LSTM}], AdapLSTM[\cite{adapLSTM}].
The TDEFSI-LONLY model employs data augmentation, the generation of additional simulated data, for training at a higher resolution than exists in their datasets. This resolves the overfitting issue associated with deep neural networks that require many parameters. This paper's application of data augmentation is to create an entirely simulated sample for training allowing the use of the real COVID-19 dataset for validation. In this study we explore the forecasting capability of the TDEFSI-LONLY model trained on SEIR generated data. We compare LONLY along with techniques found in \cite{Wangetal2020} and a dropout regularization technique \cite{Dropout} and Our new \alg{} model overcomes some of the practical parameter training issues of LONLY method. 


\def\btheta{\mathbf{\theta}}
\def\bz{\mathbf{z}}
\def\bx{\mathbf{x}}
\def\by{\mathbf{y}}
\def\bX{\mathbf{X}}
\def\bZ{\mathbf{Z}}
\def\diy{\displaystyle}
\vspace{-0.2cm}
\section{County Level Forecast Models}
\subsection{Adapted TDEFSI Model} \label{sec:TDEFSI}
\vspace{-0.2cm}
In this section, we adapt the TDEFSI model proposed in \cite{Wangetal2020} to the COVID-19 forecasting problem. The the output of each step in the TDEFSI model has a dimension for each county, allowing the network to learn spatial relationships, without requiring knowledge of specific relationships beforehand. 
Let $\by=(y_1,y_2,\ldots,y_T,\ldots)$ denote the sequence of the natural logs of daily nationwide incidence, where $y_t\in\mathbb{R}$. Let $\by^C=(y_1^C,y_2^C,\ldots,y_T^C,\ldots)$ denote the sequence of daily incidence for a particular county $C\in\mathcal{D}$, where $\lvert\mathcal{D}\lvert=K$ is the number of counties. Let $\by'_t=(y^C_t\forall C\in\mathcal{D})$. The objective is defined as predicting both nationwide and county-level incidence at day $t$, where $t = T + 1$, denoted as $\bz_t=(y_t,\by'_t)$. The TDEFSI loss function is given as:
\begin{eqnarray} \label{eq:TDEFSI_loss}
    \min\limits_{\btheta} L(\btheta)=\diy\sum\limits_{t}\|\bz_t-\widehat{\bz}_t\|_2^2+\mu \phi(\widehat{\bz}_t) +\lambda\delta(\widehat{\bz}_t),
\end{eqnarray}
where $\phi(\widehat{\bz}_t)$ and $\delta(\widehat{\bz}_t)$ are activity regularizers added to the outputs for spatial and non-negative consistency respectively, and $\lambda$ and $\mu$ are penalty parameters.
Since RNNs are not affected by the temporal scale of its input sequence, (\ref{eq:TDEFSI_loss}) does not need to be adjusted to account for the fact that COVID-19 data is updated daily as opposed to the weekly ILI data. \cite{Wangetal2020} only considered ILI data for individual states and their counties, whereas for COVID-19 we consider all counties in the nation. Min-max normalization is used on the county-level data so that $\by'\in[0,1]$. Whereas \cite{Wangetal2020} use the sum of $\by_t'$ for $y_t$, here $y_t$ is the natural log of the sum of $\by_t'$. This is done to normalize $y_t$ relative to $\by_t'$ while preserving the dependency of $y_t$ on $\by_t'$. Without this normalization, $y_t$ would be significantly larger than any  $y_t^C\in\by_t'$ for a large $K$, and would dominate the gradients used to optimize (\ref{eq:TDEFSI_loss}). This modification to $y_t$ requires that the regularization term $\phi(\widehat{\bz}_t)$ be modified to be
\begin{eqnarray} \label{eq:TLDR_regularizer}
    \phi(\widehat{\bz}_t) =\lvert \exp{(\widehat{y}_t)} - \sum_{C\in \mathcal{D}}\widehat{y}_t^C \rvert.
\end{eqnarray}
In Section \ref{sec:exp:tdefsi}, we use simulated data from the county level adapted SEIR model Section~\ref{MixSeir} to supplement available real data and analyze the adapted TDEFSI model. We provide more details on the architecture of the model in Supplementary Material. 
\subsection{\alg{}: Design Details}
\paragraph{Notations and Parameters} Denote
$n_C$ the number of counties, $n_{TF}$ the number of features represented by the LSTM backbone, $n_D$ the number of dense units, $C^{(i)}_{t}$ LSTM cell state, $n_F$ the number of forecast, and $n_{X}$ the number of additional features. In our \alg{} model we define a set of trainable parameters, $\theta_E:=\theta_{Encode}$, $\theta_{E} \in (\mathbb{R}^{2 \times n_{TF}}, \ \mathbb{R}^{n_{TF} \times n_{TF}})$
$\theta_R:=\theta_{Remember}$,$\theta_{R} \in (\mathbb{R}^{n_{TF} \times n_{TF}}, \ \mathbb{R}^{n_{TF} \times n_{TF}})$,
$\theta_F:=\theta_{Forecast}$, $\theta_{F} \in (\mathbb{R}^{n_{TF} \times n_{TF}}, \ \mathbb{R}^{n_{TF} \times n_{TF}})$, 
$\theta_{TD}:=\theta_{Time\ Distributed}$, $\theta_{TD} \in \mathbb{R}^{2 \times n_{C}}$, and for the input, middle, and output layers of the County distributed dense layer 
$\theta_{CD}:=\theta_{County \ Distributed}$, $\theta_{CD} \in (\mathbb{R}^{(n_{X}+2) \times n_{D}}$, \ $\mathbb{R}^{(n_{D}+1) \times n_{D}}, \ \mathbb{R}^{(n_{D}+1) \times 1}$).
\begin{figure}[t]
    \centering
    \includegraphics[width=0.8\columnwidth]{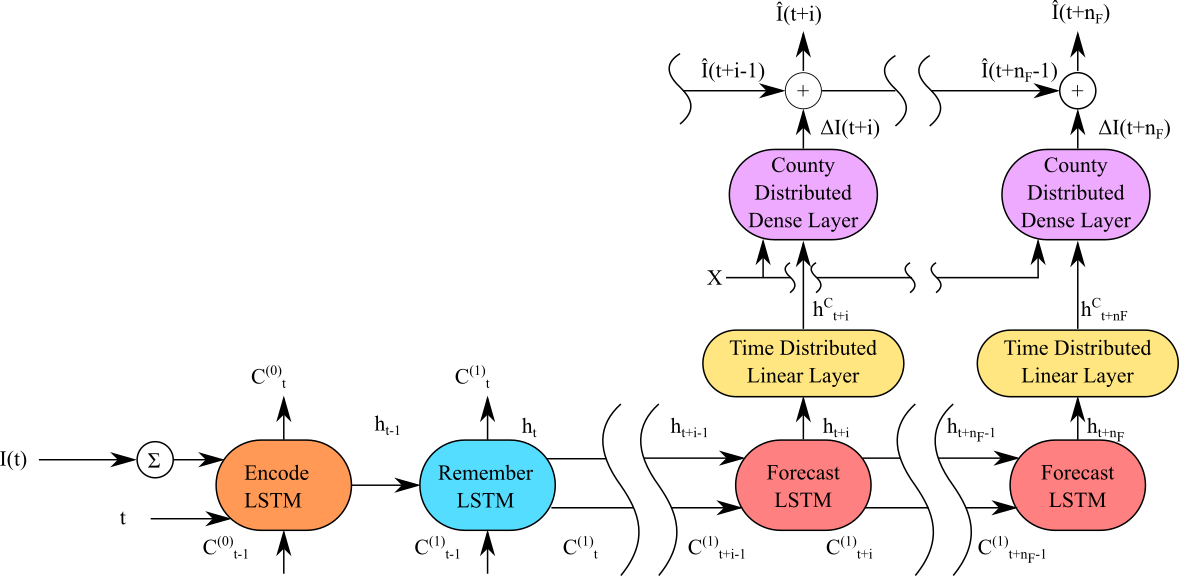}
\caption{ The \alg{} architecture consists of an LSTM backbone, time distributed layer, and a county distributed layer. Cells with shared color have shared weights. Each LSTM backbone consists of an encode cell, remember cell, and, extendable forecast cell.
}
    \label{fig:diagram}
\end{figure}
\paragraph{Description}
Our proposed \alg{} architecture, Figure \ref{fig:diagram} is designed to both to minimize the number of trainable parameters and more closely align with underlying physical phenomenon than the TDEFSI architecture. 
The goal of \alg{} is to learn to separate a high dimensional county level signal from a low dimensional national signal through a time dependent, time distributed, and county distributed hierarchy. The network is trained to encode complex national infection dynamics with a low dimensional time varying signal through an LSTM backbone when given the previous day's national confirmed cases, days elapsed since the first national recorded case, and prior day's LSTM cell state. Time invariant relationships are learned using a time distributed dense layer across all counties given the time pattern learned by the LSTM. County invariant relationships are then learned by a county distributed layer given each counties underlying time pattern and any additional county level features. Predictions from the county distributed layer are added to the previous day's county level recorded cases to predict current day's value. Using such a hierarchical representation allows us to expand the dimension of the input from the national to the county level without requiring a corresponding increase in parameters. 

The architecture consists of an LSTM backbone, time distributed layer, and a county distributed layer. The LSTM backbone consists of an encode cell, remember cell, and, extendable forecast cell as shown in Figure~\ref{fig:diagram}. The encode cell is responsible for learning to incorporate the national total recorded cases and elapsed time into the backbone time pattern, given its past batch cell state. The remember cell is responsible for remembering forecasting cell's state between batches, enabling longer term network memory without exposure to large gradients during backpropagation. The forecast cell is responsible for propagating the time pattern forward in time. Longer horizons can be forecast by repeating the forecast cell and using the preceding cell's output as input to the next. For each step in time, the time distributed layer expands the low dimensional time features learned by the forecast cell into a high dimensional county level signal. Given the county level time feature and other county level features, the county distributed layer then learns to predict each counties change in confirmed cases.

To formulate the \alg{} model, denote the function of a standard LSTM cell $\langle h_i,C_i \rangle=L(h_{i-1},C_{i-1}|\theta)$, linear dense layer $y=f(x|\theta)$, and nonlinear dense layer, $h=\sigma(y|\theta)$ where $f$ and $\sigma$ are linear and non-linear activation functions. Now let $t$ be the number of days elapsed since the first recorded national confirmed case and let $I(t) \in \mathbb{Z}^{+^{n _{C}}}$ be the vector of true county level end of day confirmed cases. The encoder and remember LSTMs are given by:
\begin{eqnarray}
\langle h_{-1},C^{(0)}_t \rangle=L\left(\left\langle \sum(I(t)),t \right\rangle, C^{(0)}_{t-1} |\theta_{E}\right),\;\; \langle h_0,C^{(1)}_t \rangle=L(h_{-1}, C^{(1)}_{t-1} |\theta_{R}),
\end{eqnarray}
where $h,C \in \mathbb{R}^{n_{TF}}$. Let $i \in 1,\ldots,n_{F}$ be the index of the day in the forecast horizon. The extendable forecasting unit consists of a forecasting LSTM vertebrae, with a time distributed linear layer {\bf \alg{} (Variant I)}, and with a combination of both time and county distributed layers {\bf \alg{} (Variant II)}. The forecasting LSTM vertebrae is given by:
\begin{eqnarray}
    \langle h_i,C^{(1)}_{t+i} \rangle=L(h_{i-1}, C^{(1)}_{t+i-1} |\theta_{F}), \;\; \;\; h_i^{C}=f(h_i|\theta_{TD}),
\end{eqnarray}
where $h_i^{C}$ is the time distributed linear layer and $h_i^{C} \in \mathbb{R}^{n_{C}}$. Let $\bX \in \mathbb{R}^{n_{X} \times n_{C}}$ be an additional county level feature matrix and let $\bZ\in \mathbb{R}^{(n_{X}+1) \times n_{C}}$ be the concatenation of $X$ with $h_i^{C}$ along the county axis, and let $j$ be the county index. The final layer predicts the change in each county's confirmed cases given it's current temporal feature through the following:
\begin{eqnarray}
    \Delta I(t+i)_j=\sigma(z_{j}|\theta_{CD}), \qquad  \Delta I(t+i) \in \mathbb{R}^{n _{C}}.
\end{eqnarray}
For $i \in 1...n_{F}$, the forecast end of day recorded cases denoted by $\widehat{I}(t+i)$ is then:
\begin{eqnarray}
    \widehat{I}(t+i)=\Delta I(t+i)+\widehat{I}(t+i-1), \;\; \hbox{where}\;\;  \hat{I}(t+1)=\Delta I(t+1)+I(t),\;\; \hbox{for}\;i=1.
\end{eqnarray}
A detailed description of \alg{} architecture is provided in the Supplementary Materials.



\paragraph{Spatial Mixing Extension to SEIR by Incorporating Inter-County Flow} \label{MixSeir}
The TDEFSI-LONLY model requires simulated training data supplemental to the JHU set used for testing. We trained using time series data generated using an inter-county population flow extended SEIR model similar to the model described in \cite{bonnasse2018epidemiological}. The full ODEs can be found in the supplementary material. For more compact form for efficient computation, let $n_C$ be the total number of counties. we denote $p_i$ total population of county $i$, $i \in \{1,2,\ldots, N_{n_C}\}$, $p_{k,i}$ Population with status $k$ in county $i$, $k \in \{S, \ E, \ I, \ R\}$, where S, E, I, R stands for {\textit Susceptible (S), Exposed (E), Infected (I), Recovered (R)} respectively. Let $x_{k,i}:=\diy p_{k,i}\big/p_i$, and $f_{i,j}$ be total population flow of people from county  $i$ to $j$ such that $f_{i,j}=f_{j,i}$ and $f_{i,i}=0$ for guaranteeing each county has no net change in population. $\mathbf{F}^{bal}=\mathbf{F} - diag(\mathbf{1}^T \mathbf{F})$. The matrix $\mathbf{F}^{bal}$ has the important property that $\sum \mathbf{F}^{bal} \times \mathbf{1} = 0$, guaranteeing the conservation laws are satisfied. Then we have the following system of ODEs
\begin{eqnarray}\begin{array}{ccl}
\diy \frac{d(\mathbf{p}_{S}(t))}{dt}= \mathbf{F}^{bal} \times \mathbf{x}_{S}(t) - \boldsymbol{\beta} \odot \mathbf{p}_{I}(t) \odot \mathbf{x}_{S}(t),\\
\diy \frac{d(\mathbf{p}_{E}(t))}{dt}= \mathbf{F}^{bal} \times \mathbf{x}_{E}(t) + \boldsymbol{\beta} \odot \mathbf{p}_{I}(t) \odot \mathbf{x}_{S}(t) - \sigma \mathbf{p}_{E}(t),\\
\diy \frac{d(\mathbf{p}_{I}(t))}{dt}= \mathbf{F}^{bal} \times \mathbf{x}_{I}(t) + \sigma \mathbf{p}_{E}(t) - \gamma \mathbf{p}_{I}(t),\;\;
\diy \frac{d(\mathbf{p}_{R}(t))}{dt}= \mathbf{F}^{bal} \times \mathbf{x}_{R}(t) + \gamma \mathbf{p}_{I}(t).
\end{array}\end{eqnarray}
$\mathbf{p}_k(t)$ is then solved for iteratively using Euler's method, see \cite{Euler}, using $\diy \mathbf{p}_k(t+h)=\mathbf{p}_k(t) + h \frac{d(\mathbf{p}_{k}(t))}{dt}$,
where $h$ is learning parameters. Next, assume that 
$f_{i,j}= \diy \min(p_i,p_j)\big/(d_{i,j}\mu_{flow})$ and  $\beta_i=\diy\rho_{i}/\mu_{spread}$,
where $d_{i,j}$ is the spatial distance between $i$ to $j$, $\rho_{i}$ is the population density of $i$, and, $\mu_{flow}$, $\mu_{spread}$ are the resistances to population flow, and infection spread respectively. The system is then controlled by the four parameters, $\mu_{flow}$, $\mu_{spread}$, $\sigma$, $\gamma$ and the initial conditions, which are estimated as follows 
\begin{eqnarray}\begin{array}{ccl}
\diy p_{E,i}(0) \sim Possion(p_{i} \lambda_E), \;\; \; p_{I,i}(0) \sim Possion(p_{i} \lambda_E \lambda_I)\\[8pt]
p_{R,i}(0)=0, \;\;\; p_{S,i}(0)=p_{i}-p_{E,i}(0)-p_{I,i}(0),  
\end{array}\end{eqnarray}
Where $\lambda_E$ and $\lambda_I$ represent the prevalence of exposure in the susceptible population, and the prevalence of infection in the exposed population at time zero, respectively. A detailed breakdown explanation of each equation is included in the Supplementary Materials.

 
\section{Experimental Results}
\paragraph{COVID-19 Dataset}
{The dataset featured is JHU confirmed cases data for US counties which is updated daily. It provided a time series of confirmed cases from 1/22/20 to 5/31/20 along with latitude and longitudinal information for each county. We use this data or a subset of this data as a single sample for the adapted TDEFSI and CLEIR-Net. Note, due to limited testing capability in the US these confirmed cases numbers are the lower bound of the actual number of infections. This paper is forecasting the number of recorded cases, rather than infections.}
 \vspace{-0.3cm}
\subsection{Evaluation of Adapted TDEFSI Model} \label{sec:exp:tdefsi}
\begin{figure}[h]
    \centering
    \subfloat[][]{\includegraphics[width=0.4\columnwidth]{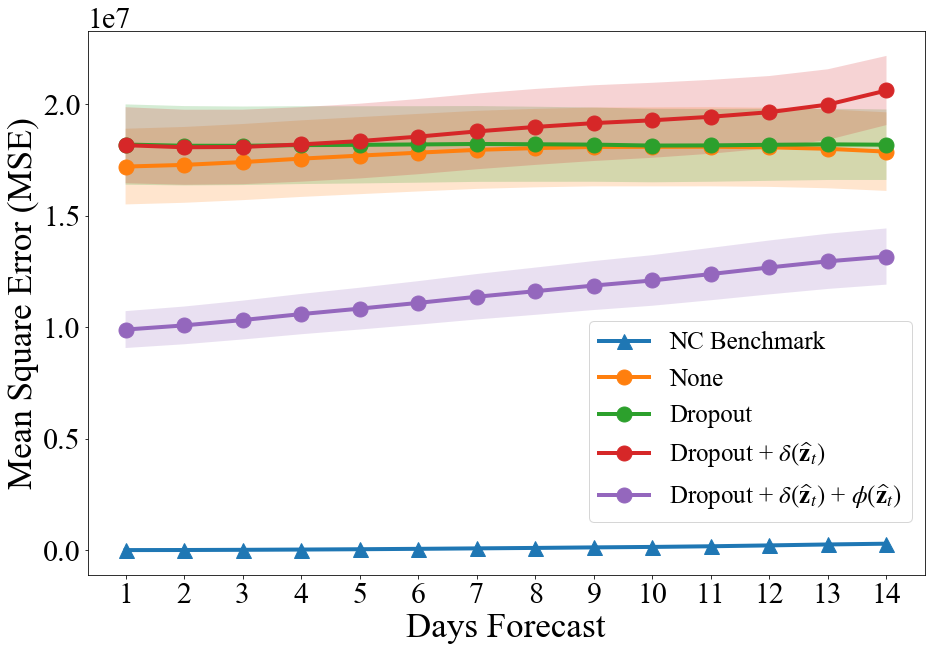}\label{fig:TDEFSI_mse_counties}}\hfill
    \subfloat[][]{\includegraphics[width=0.4\columnwidth]{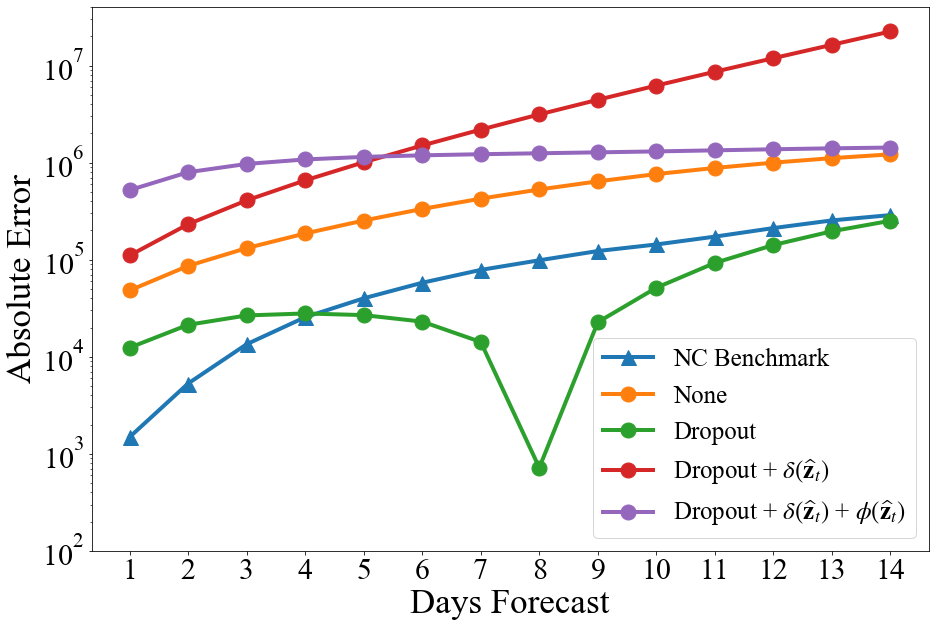}\label{fig:TDEFSI_ae_natiownide}} 
    \caption{(a) MSE with the bounds of $\pm\frac{1}{5}$ of the standard error of the two week county forecast and (b) absolute error of two week nationwide total forecast for TDEFSI models for Na{\"i}ve No-Change Benchmark\cite{}, Dropout (\cite{Dropout}), Dropout with penalty $\delta(\widehat{\bz}_t)$, and Dropout with penalty $\delta(\widehat{\bz}_t)+\phi(\hat{\bz}_t)$ in TDEFSI models.}
     \vspace{-0.2cm}
\end{figure}

 \vspace{-0.1cm}
A total of 1024 SEIR simulations as described in Section \ref{MixSeir} were run to train the adapted TDEFSI-LONLY model as described in Section \ref{sec:TDEFSI}, with an additional 16 simulations for validation. The network was trained for 300 epochs with a patience of 50. Four experiments were run: without any regularization, with dropouts, non-negative constraint regularization, and spacial consistency regularization. The final losses and MSE for each experiment are reported in Table \ref{tab:tdefsi_metrics1}. We provide the detailed list of parameters and number of hidden layers in the Supplementary Materials.\\ 
\begin{table}[t]
\vspace{-0.5cm}
\centering
\caption{Results of training the TDEFSI-LONLY model with different regularization methods.}
\begin{tabular}{lrrrrr} 
\toprule
& None & Dropout & Dropout + $\delta(\widehat{\bz}_t)$ & Dropout + $\delta(\widehat{\bz}_t)$ + $\phi(\widehat{\bz}_t)$ \\
\midrule
Train MSE & 2.57178e-5 & 5.24136e-5 & 1.54713e-4 & 2.444813e-4 \\
Valid MSE & 2.50214e-5 & 2.83442e-5 & 6.43349e-5 & 9.341941e-5 \\
Train Loss & - & - & 1.75519e-4 & 5.094754e-4 \\
Valid Loss & - & - & 8.26531e-5 & 1.906986e-4 \\
\bottomrule
\end{tabular}\label{tab:tdefsi_metrics1}
\end{table}
Figure \ref{fig:TDEFSI_mse_counties} shows the MSE of $\widehat{\by}'_t$ for each day in the forecasts for 5/18/20 to 5/31/20 based on ${y}_t$ from 1/22/20 to 5/17/20 made using the adapted TDFESI networks, and Figure \ref{fig:TDEFSI_ae_natiownide} shows the absolute error of $\widehat{y}_t$ for each day in that forecast.
\subsection{\alg{} Forecast Results} \label{sec:cleirn-forecast}
The model is trained with features from the 1/22/20 to 5/2/20 period, targets from 1/23/20 to 5/16/20 with validation features from 5/3/20 to 5/4/20 and targets from 5/3/20 to 5/17/20. The trained model is tested by making a single 14 day forecast over 5/18/20 to 5/31/20 for all counties. The county level temporal patterns are supplemented with standard normalized county level $latitude$, $longitude$, $population$, and $population \ density$ \cite{killeen2020countylevel}, as well as $\log(population)$ and $\log(population \ density)$ prior to the application of the county distributed dense layer. 
We represent $n_{TF}$ features in the LSTM backbone and use 3 layers of $n_D$ units in the county distributed branch taking $\sigma$ to be the ReLU activation. Both training and inference use batch size of 1, with batches taken in sequential order, and previous encoding cell state shared with the next batch. Each batch uses the vector of the previous end of day's recorded confirmed cases to forecast a target matrix of size $(n_{C}, n_{F})$. Dropout is applied to the targets at a rate of $0.25$ during training to mitigate overfitting.

\begin{figure}[H]
    \centering
    \subfloat[][]{  
    \includegraphics[width=0.44\textwidth]{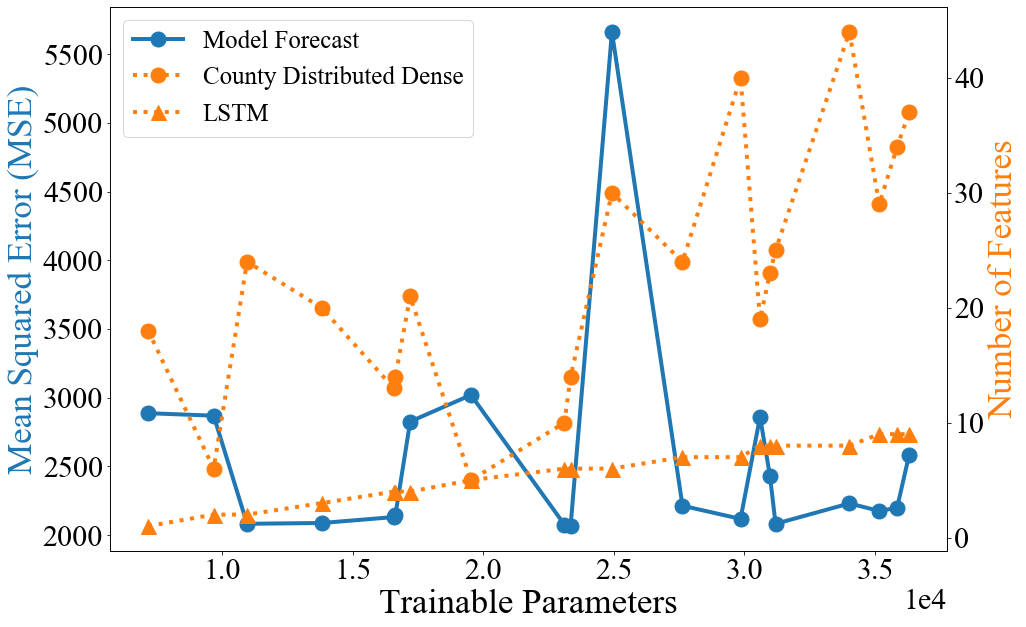}\label{fig:mint_snacs_runtime}}\hfill
    \subfloat[][]{ \includegraphics[width=0.40\textwidth]{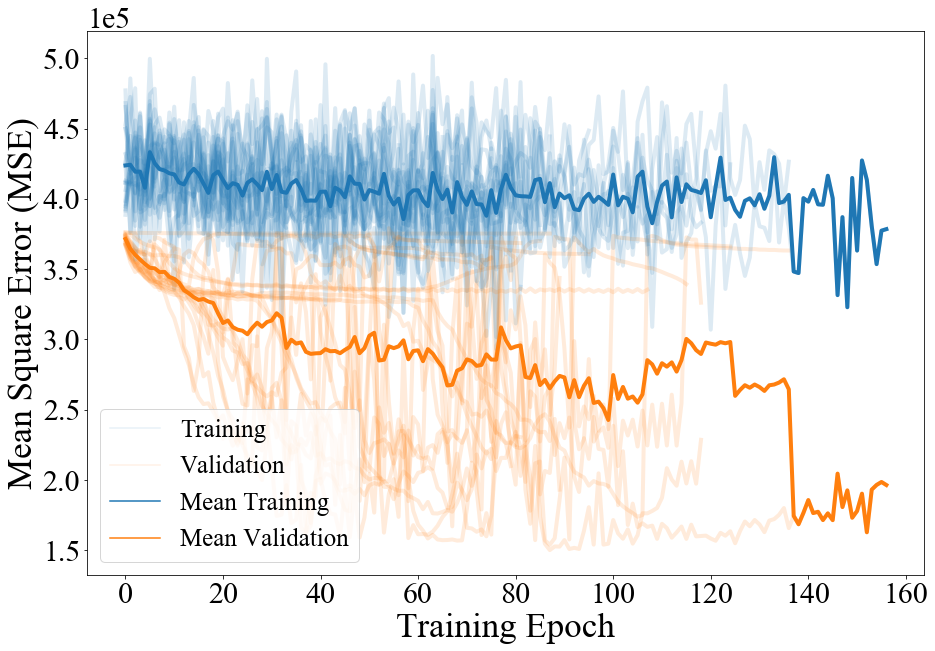}\label{fig:snacs_runtime}}
    \caption{(a) The relationship between the number of parameters in various configurations of CLEIR-Net models and both the Mean Squared Error (MSE) of forecasts using those models, as well as the the number of features in the dense and the LSTM layers of those models. Note that the number of trainable parameters is scaled using scientific notation as notated in the bottom right corner of the plot. (b) Individual and mean training and validation curves for CLEIR-Net models.}
    \label{fig:Benchmarks}
     \vspace{-0.5cm}
     \setlength{\belowcaptionskip}{10pt}
\end{figure}

We perform a capacity study \ref{CLEIR Ensemble Sumary} by randomly sampling 20 configurations from $\{n_D,n_{TF} \in \mathbb{Z} | \quad 10 \leq n_D \geq 50, \ 1 \leq n_{TF} \geq 5\}$ with equal probability. While there is no clear trend in forecast mean squared error; we observe that a network with $n_{TF}=5, \ n_D=20$ is reproducible and lightweight. Anecdotally, training proceeds best on the edge of instability, encouraging annealing of the network towards more stable configurations. Different size networks learn in different regimes, and capacity must be tuned in conjunction with dropout rate. We note that sharing cell state between sequential batches is essential for network convergence to an effective forecasting configuration. Using both the time distributed and county distributed dense layers is substantially more effective than using only a time distributed, or a time distributed, county distributed layer. Additionally, predictions from all 20 model configurations are ensembled by averaging and scored over the forecast horizon. The ensembled predictions perform better than any single model's predictions, even in the presence of poor scoring component models. Due to the lightweight nature of of the architecture, ensembled predictions are easily obtained, mitigating the instability inherent in dropout and common in RNN architectures. 

\begin{figure}[H]
    \centering
    \includegraphics[width=0.8\textwidth]{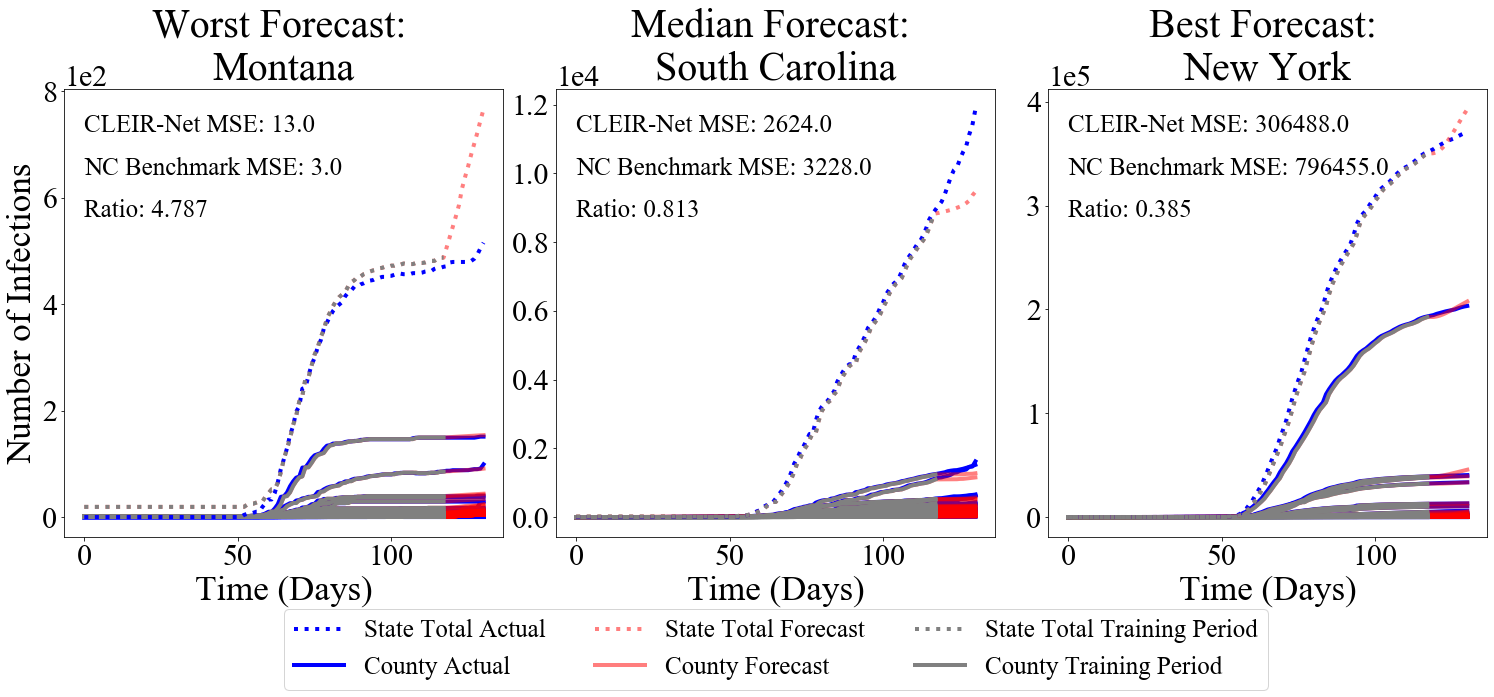}
    \caption{The best, worst, and median state forecasts made using CLEIR-Net Model, based on the ratio of the MSE of the forecast to the MSE of the na{\"i}ve no change benchmark.}
    \label{"BestWorstMedian"}
\end{figure}

\begin{figure}[h]
\minipagett{0.4}{0.5}{
\includegraphics[width=1.2\textwidth]{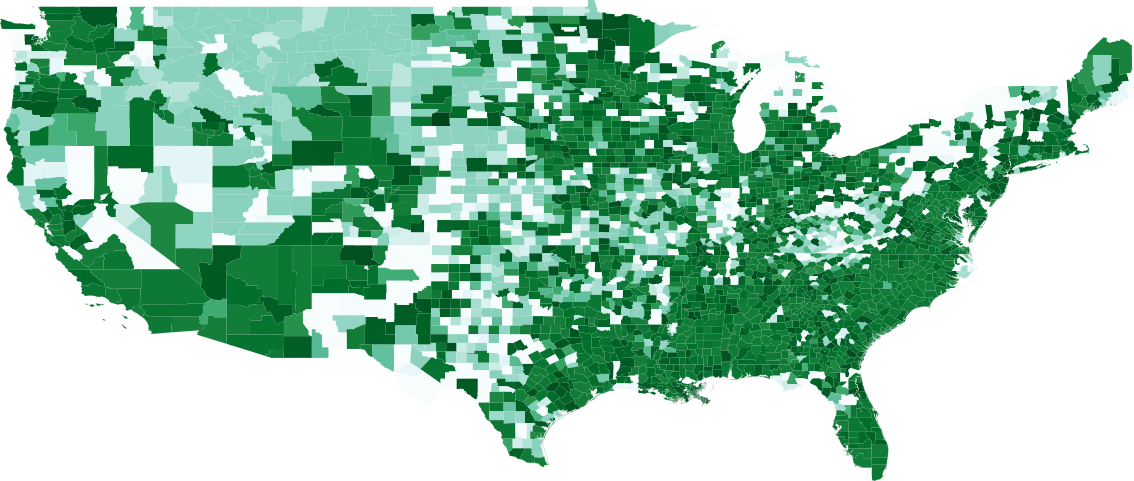}
}{
\caption{A map showing the mean square error for the 48 contiguous states. Darker values represent a lower MSE. Here we can see the trends where the model performs the worst for low connected counties. This is most easily observed in the trend seen through the very rural mid-west states.
}
\label{fig:data stream 2}
}
\end{figure}

\begin{figure}
    \centering
    \includegraphics[width=0.9\textwidth]{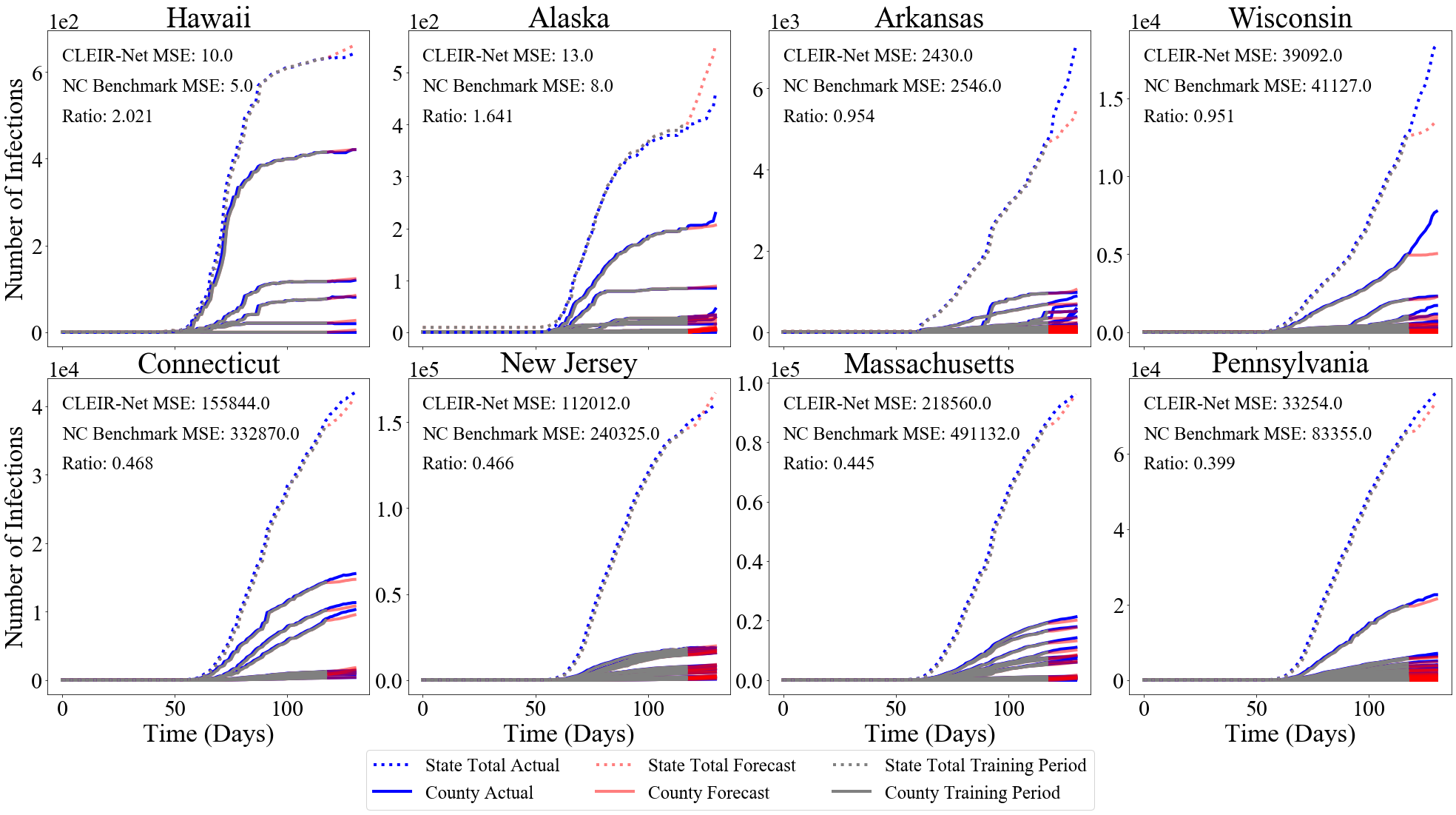}
    \caption{The second through fourth worst, and the fourth through second best forecasts made using CLEIR-Net Model, based on the ratio of the MSE of the forecast to the MSE of the na{\"i}ve no change benchmark.}
    \label{"BestWorst5"}
\end{figure}

The best and worst performing model forecasts are collected in Figure \ref{"BestWorstMedian"}. Figure \ref{"BestWorst5"} shows the second through fourth worst and the forth through second best forecasts made using \alg{} model. 
Additional forecasts are provided in the Supplementary Material. 
Best predicted by CLEIR-Net are large, connected, counties. In particular, the top five best scoring state forecasts were the New York and four adjacent states: Pennsylvania, Massachusetts, New Jersey, and Connecticut. Worst predicted are small or isolated counties, such as Hawaii, and Montana. Also, poorly predicted are states with peculiar dynamics relative to the rest of the nation. For example, the five best predicted states exhibit the early stages of curve flattening, while Arkansas, and Wisconsin, the fourth and fifth worst predictions have an accelerating number of cases. Figure \ref{fig:data stream 2} illustrates this via a map showing the MSE for the contiguous states. 

A summary of the forecast performance and size of all the models explored in this paper is presented in Table \ref{CLEIR Ensemble Sumary}. All methods were used to forecast for 5/18/20 to 5/31/20. We observe that \alg{} requires significantly low numbers of parameter to train versus the adapted TDFESI model. 

\begin{table}[H]
\centering
\caption{A comparison of the number of parameters and the MSE of forecasts made with the benchmarks, CLEIR-Net variants, and adapted TDFESI models.}
\begin{tabular}{llrrr}
\toprule
 Model & Variant & MSE &    Total Parameters \\
\midrule
\multirow{1}{*}{Benchmark}& Na{\"i}ve No Change     &  108276  & -  \\
\hline
\multirow{4}{*}{CLEIR-Net} & Average Ensemble of 20      &   69870  & -\\
& Best Scoring   &   70905  & 10945 \\
& Median Scoring &   75707  & 35847 \\
& Worst Scoring &  188677   & 24923 \\
\hline
\multirow{4}{*}{TDFESI} & None & 17798679  & 1038405 \\
& Dropout & 18178904  & 1038405 \\
& Dropout + $\delta(\widehat{\mathbf{z}}_t)$ & 18950460  & 1038405 \\
& Dropout + $\delta(\widehat{\mathbf{z}}_t)$ + $\phi(\widehat{\mathbf{z}}_t)$ & 11495507  & 1038405 \\
\bottomrule
\end{tabular}
\label{CLEIR Ensemble Sumary}
\end{table}

\section{Discussion}
As expected, adding regularization terms to the loss function of the TDEFSI model increased the final loss and MSE, see Table 1. However, regularization did improve the generalization of the model, making it less likely to overfit the simulated training data. As seen in Figure \ref{fig:TDEFSI_mse_counties}, combining the regularization methods improved the model's county level forecasting performance, but underperforms the na{\"i}ve no-change benchmark. The dropout method was the only regularization technique to improve the national forecasting performance, even outperforming the na{\"i}ve no-change benchmark when forecasting more than four days out. While the TDEFSI model is viable for ILI forecasts as shown by \cite{Wangetal2020}, the minor adaptations to the TDEFSI model tested here were insufficient to make the model a viable method for forecasting COVID-19 confirmed case data. More extreme adaptations or altogether different architectures, such as CLEIR-Net, must be considered.

%
There are many patterns superimposed on top of underlying pandemic dynamics, such as state, national, and temporal variations in policies for social distancing. This requires the careful balancing of capacity to fit complex observations without overfitting. 
CLEIR-Net is lightweight and scalable in terms of features and targets due to its hierarchical construction. While the number of features used by the LSTM backbone should be kept low since the time distributed has weights of shape $(n_{TF}+1) \times n_{C}$ future constructions might rethink this step. Further, there is likely room to learn stronger signals from appropriate features with existing capacity. Naturally, we should seek the smallest set of features for the neural network with the most explanatory power. This could be through the addition of mobility or policy data.


\paragraph{Limitations and Future Work}
There are several experiments we wish to see explored. These models national forecasts could be improved through optimizing the SEIR model parameters. The sampling ranges of SEIR parameters could be extended to allow more diversity of simulation to ensure we do not restrict the model's ability to capture all of the observed patterns. 
Future work on CLEIR-Net could consider improving the realism of the model and the connection of the network with underlying dynamics, such as by incorporating SEIR mechanics. Additional county level features might be incorporated at the county distributed layer. 
This same work could be done with the models trained on JHU global confirmed cases. In this case the countries would take the place of the counties in the current models and a world forecast would take the place of the national forecast in the current models. 
Additional work could be done to apply these methods to other measures of epidemiological progression, such as hospitalizations, death, and recovery rate.

\section{Broader Impact}
Two critical steps in the medical resource supply chain are procurement and deployment. For each, decision makers require their own specialized forecasting tools. The power of the proposed CLEIR-Net architecture is it's ability to forecast over a large area concurrently at a higher resolution and for less computational expense than existing models. The model can give healthcare administrators in charge of medical resource allocation a two week lead time to optimize nationwide resource deployment to areas with the highest need and save lives. Meanwhile, many existing models, which are simpler and less computationally expensive, are sufficient to forecast single dimension national or regional pandemic dynamics and can reliably do so over longer horizons than two weeks to the benefit of system administrators in charge of procurement who must predict national and regional resource demand.

\section{Acknowledgement}
The authors would like to thank John Brindley, Mohsen Alizadeh Noghani, and Jens Early Hansen for their contribution to primary results established during the course “COS598 - Machine Learning” at University of Maine.

\bibliographystyle{plainnat}
\bibliography{refs}









\newpage
\renewcommand{\thesection}{1}
\def\diy{\displaystyle}
\begin{center}
    {\Large \bf Supplementary Materials}
\end{center}
\section{Spatial mixing Extension to SEIR}\label{SM:SERI}

\subsection{The SEIR model}

The SEIR model assigns one of four status to a proportion of the total population in a space with constant population. It then builds a set of ordinary differential equations (ODEs) which describe rate of change of each status population. We use a version of the SEIR model that neglects birth and death effects and assumes a lack of any vaccination though it assumes recovered patients gain permanent resistance. The model is identical to one found in \cite{R_hubei} which is recreated below. 

\begin{eqnarray}\begin{array}{l}
\diy \frac{d(p_{S,i}(t))}{dt}= \diy - \beta_i p_{I,i}(t)  x_{S,i}(t), \;\;\;
\diy \frac{d(p_{E,i}(t))}{dt}= \beta_i p_{I,i}(t)  x_{S,i}(t) - \sigma p_{E,i}(t),\\
\diy\frac{d(p_{I,i}(t))}{dt}= \sigma p_{E,i}(t) - \gamma p_{I,i}(t),\;\;\;
\diy\frac{d(p_{R,i}(t))}{dt}= \gamma p_{I,i}(t),\;\;\;
\diy S + E + I + R = N.
\end{array}\end{eqnarray}

We denote $p_i$ to be the total population of county $i$, and $p_{k,i}$ to be the population with status $k$ in county $i$, where $k \in \{S, \ E, \ I, \ R\}$. $x_{k,i}:=\diy p_{k,i}\big/p_i$ is the proportion of the population with status $k$. The parameters, $\beta$, $\sigma$, and $\gamma$, along with initial conditions govern the dynamics of the epidemic. This model does not account for any interaction between the given county's population and all other U.S. counties.

\subsection{County Mixing SEIR Derivation}
A conservation law for the population of the county requires the rate of population accumulation to equal the rate of population inflow minus the rate of population outflow plus the rate of population generation. The population generation term is given by our SEIR equations. We assume no change in the county population requiring the population flow and generation terms to add to zero. Additionally, intuitively we know the SEIR equations which govern the transition of status should not induce population accumulation or depreciation. This requires our population flow terms to balance.

We decided on population mixing terms that would account for the change in a status' proportion of the population without changing the total county population. To do this, the inflow term for given $k$ population in county $i$ is a foreign county's population flow into county $i$ multiplied by the foreign county's proportion of population with status $k$ summed over all foreign counties. Symmetrically, the outflow term is the same population flow variable, but this time multiplied by proportion of population with status $k$ in county $i$ summed over each foreign county. The multiplication of the proportions, which must sum to one, and the flow terms, which are equivalent across ODEs for a county pair, ensure the county population does not change.

Let $f_{i,j}$ be the total population flow of people from county  $i$ to $j$ such that $f_{i,j}=f_{j,i}$ and $f_{i,i}=0$. These conditions guarantee each county has no net change in population. The derivatives of $p_{k,i}$, $k \in \{S,E,I,R\}$ can be written as

\begin{eqnarray}\begin{array}{l}
\diy \frac{d(p_{S,i}(t))}{dt}= \diy \sum_{j} f_{i,j} x_{S,j}- x_{S,i} \sum_{j} f_{j,i} - \beta_i p_{I,i}(t)  x_{S,i}(t), \\
\diy \frac{d(p_{E,i}(t))}{dt}= \sum_{j} f_{i,j} x_{E,j} - x_{E,i} \sum_{j} f_{j,i} + \beta_i p_{I,i}(t)  x_{S,i}(t) - \sigma p_{E,i}(t),\\
\diy\frac{d(p_{I,i}(t))}{dt}= \sum_{j} f_{i,j} x_{I,j} - x_{I,i} \sum_{j} f_{j,i} + \sigma p_{E,i}(t) - \gamma p_{I,i}(t),\\
\diy\frac{d(p_{R,i}(t))}{dt}= \sum_{j} f_{i,j} x_{R,j} - x_{R,i} \sum_{j} f_{j,i} + \gamma p_{I,i}(t).
\end{array}\end{eqnarray}

We can write this system of equations more concisely if we vectorize each equation. This can be done by creating a flow matrix, $\mathbf{F}$. The elements are made up of the aforementioned $f_{i,j}$ terms. Again, using the constraints $f_{i,j}=f_{j,i}$ and $f_{i,i}=0$ we create a symmetric flow matrix with a diagonal of zeros. Here, the $i$th column represents population outflows from county $i$ and the $i$th row represents population inflows to county $i$. Lastly, we need our matrix to satisfy our constant population constraint. In other words, we want the diagonals of the matrix to be the negative sum of the other elements in their row. This can be done by subtracting the diagonal of one transpose $\mathbf{F}=\left\{f_{ij}\right\}_{ij}$, written as  $\mathbf{F}^{bal}=\mathbf{F} - diag(\mathbf{1}^T \mathbf{F})$. This matrix exhibits the property $\sum \mathbf{F}^{bal} \times \mathbf{1} = 0$, guaranteeing the conservation laws are satisfied.



\renewcommand{\thesection}{2}
\section{Discussion on \alg{} Model}


\subsection{CLEIR-Net Training}

The network is trained over a sequence of inputs where the target is to forecast the number of recorded infections for each county for each of the next $n_{F}$ days in the future, given the days elapsed since the first national recorded infection, $t$, the vector of the current day's county level recorded infections, $I(t)$, and, crucially, the time invariant county level features, latitude, longitude, population, population density, log of population and population density. A batch size of one is used where each sample consists of a single given day of the pandemic and the corresponding targets are the days in it's forecast horizon. That is, a sequence of overlapping samples, taken in sequential order. Additionally, each batch's encoder and remember cells use the cell and hidden states from the previous batch's encoder and remember cells as their initial states and provide their output states to the next sample's corresponding cells. Sharing states between batches allows the LSTM cell to effectively utilize its memory property over the entire sequence but limit gradient exposure to short sequences. An initial state of zero is used for the first sample in the sequence.

\subsection{CLEIR-Net Forecasting}

After training, learned weights are used to forecast future infection trajectories. Since the model learns to use states from past batches, effective inference requires the model be initialized, post learning and prior to forecasting, by making sequential predictions over the entire training data.

\subsection{CLEIR-Net Components}


The network's basic definitions, inputs, targets, and outputs for a single batch are summarized here.

\textbf{Definitions}
\begin{itemize}
    \item $n_F:=n_{Forecast}$ Number of days in the forecast horizon
    \item $n_{TF}:=n_{Time \ Features}$ Number of time features to model in the LSTM backbone
    \item $n_C:=n_{Counties}$ Number of counties in the prediction space
    \item $n_X:=n_{County Level Features}$ Number of additional county level features used by the county distributed dense layer
    \item $n_D:=n_{Dense \ Units}$ Number of units in the county distributed dense layer 
\end{itemize}

\textbf{Inputs: }
\begin{itemize}
\item $t$: Time elapsed in days since first nationally recorded infection

\item $I(t)$: Current day's county level infections
\item $X$: Time invariant county level features
\item $C^{(E)}_{t-1}$: State of the previous batch's encoder cell
\item $h^{(E)}_{t-1}$: State of the current batch's encoder cell
\item $C^{(R)}_{t-1}$: State of the previous batch's remember cell
\item $h^{(R)}_{t-1}$: State of the current batch's remember cell
\end{itemize}
\newpage
\textbf{Targets:} 

The goal of the network is to forecast the end of day recorded infections for each US county for each day in the forecast horizon, therefore our targets are the following.
\begin{itemize}
\item $I(t+i) \quad \forall \ i \in 1...n_{F}$: Actual end of day recorded county level infections for the $i^{th}$ day in the forecast horizon.
\end{itemize}

\textbf{Outputs:} The corresponding model predicted infections are then denoted by the following.

\begin{itemize}
\item $\hat{I}(t+i) \quad \forall \ i \in 1...n_{F}$: Predicted end of day county level infections for the $i^{th}$ day in the forecast horizon. 
\end{itemize}

Additionally, each batch share's the hidden and cell states of it's encoder and remember LSTMs with the following batch.

\begin{itemize}
\item $C^{(E)}_{t}$: Cell state of the current batch's encoder cell
\item $h^{(E)}_{t-1}$: Hidden state of the current batch's encoder cell
\item $C^{(R)}_{t}$: Cell state of the current batch's remember cell
\item $h^{(R)}_{t-1}$: Hidden state of the current batch's remember cell
\end{itemize}

The LSTM backbone of the model consists of an encoder cell and remember cell which translate the national total recorded infections and days elapsed into a low dimensional time varying pattern, and a repeatable forecast cell which propagates the time pattern into the future.

\textbf{Encoder Cell:} 

\begin{itemize}

\item \textbf{Purpose:} The encoder cell, given the previous batch's encoder state, transforms the total national recorded infections, $\sum I(t)$, and days elapsed since the first nationally recorded infection into a low dimensional time varying pattern.  
    
\item \textbf{Inputs:} $t$, $\sum I(t)$, $C^{(E)}_{t-1}$, $h^{(E)}_{t-1}$
    
\item \textbf{Outputs:} $C^{(E)}_{t}$, $h^{(E)}_{t}$
    
\item \textbf{Parameters:} $\theta_E:=\theta_{Encode}$, $\theta_{E} \in (\mathbb{R}^{2 \times n_{TF}}, \ \mathbb{R}^{n_{TF} \times n_{TF}})$

\end{itemize}

\textbf{Remember Cell:}
\begin{itemize}   
\item \textbf{Purpose:} The remember cell, given the context from it's previous state, and the output representation from the encoder, learns to remember the current state of the LSTM backbone used by the repeated forecast cell.
    
\item \textbf{Inputs:} $C^{(R)}_{t-1}$, $h^{(R)}_{t-1}$, $h^{(E)}_{t}$
    
\item \textbf{Outputs:} $C^{(R)}_{t}$, $h^{(R)}_{t}$
    
\item \textbf{Parameters:} $\theta_R:=\theta_{Remember}$,$\theta_{R} \in (\mathbb{R}^{n_{TF} \times n_{TF}}, \ \mathbb{R}^{n_{TF} \times n_{TF}})$,

\end{itemize}

\textbf{Forecast Cell:}
\begin{itemize}
\item \textbf{Purpose:} The forecast cell is similar to the remember cell in that its function is to model the underlying low dimensional time pattern. However, while the remember cell incorporates information both from the encoder cell and its own previous state, the forecast cell uses only the LSTM backbone state. For the first day in the forecast, the cell propagates the current state from the remember cell into the future. To achieve multiple days in the forecast horizon, the forecast cell is repeated, with each consecutive cell after the first taking the state from the previous cell as input. 
    
\item \textbf{Inputs:} $C^{(F)}_{t+i-1}$, $h^{(F)}_{t+i-1}$ when $i>1$, $C^{(R)}_{t}$, $h^{(R)}_{t}$ otherwise
    
\item\textbf{Outputs:} $C^{(F)}_{t+i}$, $h^{(F)}_{t+i}$
    
\item \textbf{Parameters:} $\theta_F:=\theta_{Forecast}$, $\theta_{F} \in \mathbb{R}^{n_{TF} \times n_{TF}}$, 

\end{itemize}

The time and county distributed layers are used to enforce a hierarchical framework of shared factors. The time distributed layer uses the same parameters across all time steps to learn a time pattern underlying each county, given the low dimensional patterns from the backbone. Then, given the underlying time pattern for each county, the county distributed layer predicts the day to day change in infections using the same set of explanatory factors shared by each county.

\textbf{Time Distributed Layer: }
\begin{itemize}

\item \textbf{Purpose:} If we consider the low dimensional time pattern learned by the LSTM backbone to be a mixed signal representing the underlying national dynamics, then the goal of the time distributed linear layer is to separate the national time pattern into its county level components, providing a county level time dependent condition on which the county distributed layer is applied.

\item \textbf{Inputs:} $h^{(F)}_{t+i}$
    
\item \textbf{Outputs:} $h^{(C)}_{t+i}$
    
\item \textbf{Parameters:} $\theta_{TD}:=\theta_{Time\ Distributed}$, $\theta_{TD} \in \mathbb{R}^{2 \times n_{C}}$

\end{itemize}

\textbf{County Distributed Layer:}
\begin{itemize}
\item \textbf{Purpose:} Given descriptive factors comparable across counties, latitude, longitude, population, population density, and log population and population density, and the county level time feature learned by the time distributed layer, the county distributed layer predicts the county level daily changes in infections for all counties, $\Delta I(t)$.

\item \textbf{Inputs:} $h^{(C)}_{t+i}$, $X$
    
\item \textbf{Outputs:} $\Delta I (t+i)$
    
\item \textbf{Parameters:} $\theta_{CD}:=\theta_{County \ Distributed}$, $\theta_{CD} \in (\mathbb{R}^{(n_{X}+2) \times n_{D}}$, \ $\mathbb{R}^{(n_{D}+1) \times n_{D}}, \ \mathbb{R}^{(n_{D}+1) \times 1}$) (For the input, hidden, and output layers of the County distributed dense layer.)

\end{itemize}

\textbf{Final Prediction:}
\begin{itemize}
    \item \textbf{Purpose:} The predicted changes in infection, $\Delta I(t+i-1)$, are added to the previous day's predicted county level infections, $\hat I(t+i-1)$, to determine the current day's predicted infections, $\hat I(t+i)$.
    \item \textbf{Inputs:}  $\Delta I(t+i-1)$, and $\hat I(t+i-1)$ when $i>1$, or $I(t)$ otherwise
    \item \textbf{Outputs:} $\hat I(t+i)$
    \item \textbf{Parameters} None
\end{itemize}

\renewcommand{\thesection}{3}
\section{Experimental Setup}
\subsection{\alg{} Forecast}
Network parameters are optimized by minimizing the weighted mean squared error (MSE) between the forecast and target matrices with NAdam with a learning rate of 0.001. Each sample is weighted by a factor of $w_{i,j}=(\log (Population_j+1) \log (i+1))^{-1}$, where $i$ is the target number of days ahead in the forecast, and $j$ is the county. Training is stopped when the validation loss is unimproved for 30 epochs at which point weights from the best performing epoch are used for inference. Random dropout is applied to the targets during training at a rate of 0.25 to mitigate overfitting to any particular county's patterns. Light L1 and L2 regularization of 0.00005 are applied to the kernel and bias weights of all dense layers and to all recurrent weights to improve conditioning of the gradient. Forecasting is performed using learned parameters by resetting cell states and making predictions with a batch size of one in sequential order over all available days in the input range; ensuring the memory property is active during inference. Predictions from the last available day are taken as the forecast. \\
Prior to training, 4 counties from New York are removed from the data. Bronx, Kings, Queens, and Richmond counties have no recorded cases in the JHU data and are believed to be aggregated into the New York City totals.

Table \ref{WithandWithoutCountyDistribution} compares the forecast performance of the \alg{} architecture both with the county distributed layer (Variant II) and without (Variant I). In both cases the time distributed layer is used to transform the national time pattern into the county level time patterns. Variant I uses the time distributed layer to directly predict the county level daily recorded infection changes while Variant II uses the county distributed layer to make further refinements to the output of the time distributed layer to predict the the county level daily recorded infection changes. In both cases, predicted changes in infections are added to the previous days predicted infections for each day forecast into the future and the cost is computed between predicted and recorded infections over the entire forecast. We train both variants using both mean squared logarithmic error and mean squared error to forecast a 7 day horizon. Variant I slightly outperforms the na{\"i}ve no-change benchmark by all measures. When trained using the mean squared error objective, Variant II significantly outperforms the mean squared error of Variant I and the na{\"i}ve no-change benchmark. This yields a more meaningful predicted national change in cases but at the expense of the mean squared logarithmic error, reducing accuracy in counties with lower recorded infections.

\begin{table}[h]

\centering
\caption{Forecast metrics for both CLEIR-Net variants trained with MSE and MSLE loss functions, and the na{\"i}ve no-change benchmark. The Mean Squared Error (MSE), Mean Squared Logarithmic Error (MSLE), Mean Absolute Error (MAE), and Predicted Confirmed Case Increase (PCCI) of the forecasts using each method.}
\begin{tabular}{lrrrrr}
\toprule
{} & {} & \multicolumn{2}{c}{Variant I Objective} & \multicolumn{2}{c}{Variant II Objective} \\
Metric &  7 Day Na{\"i}ve Benchmark & MSLE & MSE & MSLE & MSE \\
\midrule
MSE     & 34190.0000    & 34095.0000    & 33727.0000    & 34102.000 & 14191.000 \\
MSLE    & 0.0366        & 0.0337        & 0.0342        & 0.033     & 1.149 \\
MAE     & 28.9100       & 28.4600       & 27.3900       & 28.400    & 30.850 \\
PCCI     & 0.0000        & 4279.0000     & 10024.0000    & 4851.000  & 140971.000 \\
\bottomrule
\end{tabular}
\label{WithandWithoutCountyDistribution}
\end{table}


\subsection{Adapted TDEFSI Experiments}

The architecture of the model used in the TDEFSI experiments in Section \ref{sec:exp:tdefsi} is adapted specifically from the TDEFSI-LONLY model. This consists of $k$ LSTM layers, each with a latent dimension of $H^{(i)}$, followed by a dense layer with $H$ outputs, and a final dense layer with $K+1$ outputs. The input sequence to the network is $\by$, and the output sequence is $\widehat{\bz}$.
All TDEFSI-LONLY experiments used the parameters $k = 2$, $H^{(i)} = 128$, $H = 256$, $\lambda = 0.01$, $\mu = 0.0001$ when relevant, which \cite{Wangetal2020} found to be optimal for their data. 


\subsection{County Dependency}

Through determining the dependency between confirmed case of counties we can predict which counties are likely to be accurately forecasted by \alg{}. The model, seeking county level and national level trends, fits itself to counties which are more inter-connected with other counties since their influence on each other induces overall trends. Our approach is to determine a non-linear dependency between counties by calculating the mutual information (MI), between counties to determine the connection between counties.

To estimate the MI between two counties, $C$ and $C'$, denoted $\rho(C,C'):=\widehat{MI}(C,C')$, we use the hash-based method, proposed in \cite{mortezaetal2018}. 
However, it is prohibitively expensive to compute the MI between every county pair. Furthermore, intuitively there is lower dependency between counties that are far apart. Therefore, we compute MI between a county and its neighbors and average over all neighbors. We use the US Census Bureau's county adjacency data to determine neighbors. 
Our approach is described in Algorithm 1. 

\begin{algorithm}[h]
\SetAlgoLined
\SetKwInOut{Input}{Input}
\SetKwInOut{Output}{Output}
\Input{Time series confirmed case data for all counties, 
      Neighbor data for all counties}
\Output{Average County Dependency $\overline{\rho}$}
\ForEach{County}{
    \ForEach{Neighbor}{
        $\rho$ = $\widehat{MI}$ (County, Neighbor)
    }
   $\overline{\rho}$ = Average $\rho$\\
}
\ForEach{$\overline{\rho}$}{
   $\overline{\rho}$ = $(\overline{\rho} - min(\overline{\rho})) / (min(\overline{\rho}) - max(\overline{\rho}))$\\
}
\caption{Confirmed Case Dependency between Counties }
\end{algorithm}

\begin{figure}[h]
\centering
\includegraphics[width=0.9\textwidth]{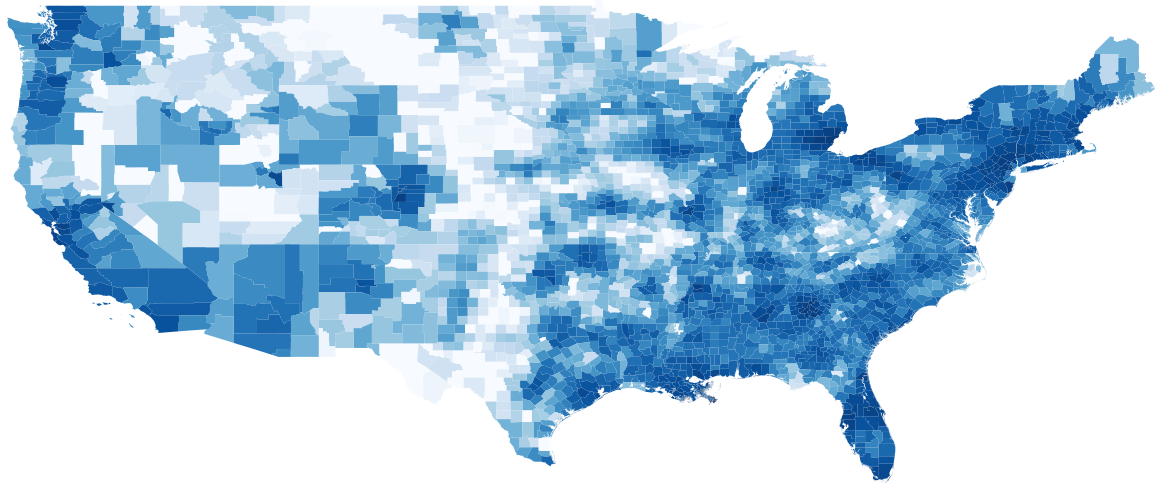}
\caption{A map showing the dependency values, $\overline{\rho}$, for the 48 contiguous states. Darker values represent a higher average dependency. As expected we observe that the more rural counties, especially as seen in the very rural mid-west states, have low inter-county dependency. We also observe a pattern of highly connected counties acting as centralized hubs, with higher connectivity of their neighbors, dissipating outwards.}
\vspace{-0.6cm}
\label{fig:dependency}
\end{figure}
\begin{figure}[H]
\centering
\includegraphics[width=0.9\textwidth]{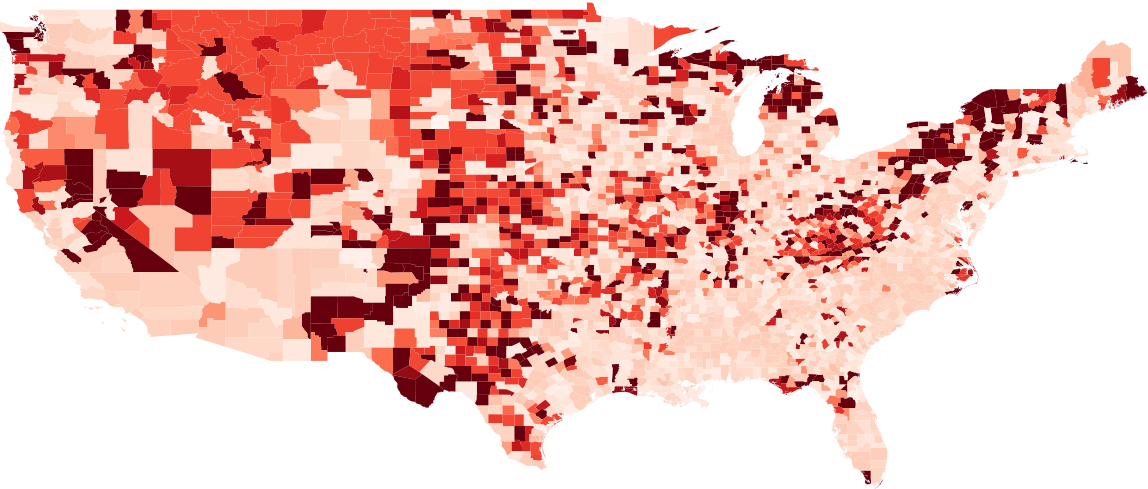}
\caption{A map showing the mean square error for the 48 contiguous states. Darker values represent a higher MSE. Note that this figure is the inverse of Figure 5 to highlight the inverse correlation between this figure and Figure \ref{fig:dependency}}
\vspace{-0.6cm}
\label{fig:error}
\end{figure}

The dependency values and error for the 48 contiguous states are shown in Figures \ref{fig:dependency} and \ref{fig:error},  respectively. We observe the correlation between the dependency values and error of counties after training. We show that the inverse correlation between dependency and error is significantly high. 


Note that by calculating the county dependency we can get a quick estimate of which counties won't be accurately forcasted by the \alg{} model before having to train it. This information is crucial to have as soon as possible because if we want to delegate resources based on the results of the model we have to know and disclose the limitations so that they can be planned around ahead of time. There are further potential applications for the dependency values as well. Notable to this work is potentially estimating flow rates between counties in the spatial mixing extension to SEIR outlined in Section \ref{SM:SERI}.

\subsection{Additional Figures for Results}

Figure \ref{"fig:all_states"} shows the CLEIR-Net forecasts for all states and the District of Columbia, except those already shown in Section \ref{sec:cleirn-forecast}, Figure \ref{"BestWorst5"}.


\begin{figure}[H]
    \centering
    \subfloat[]{
    \includegraphics[width=0.9\textwidth]{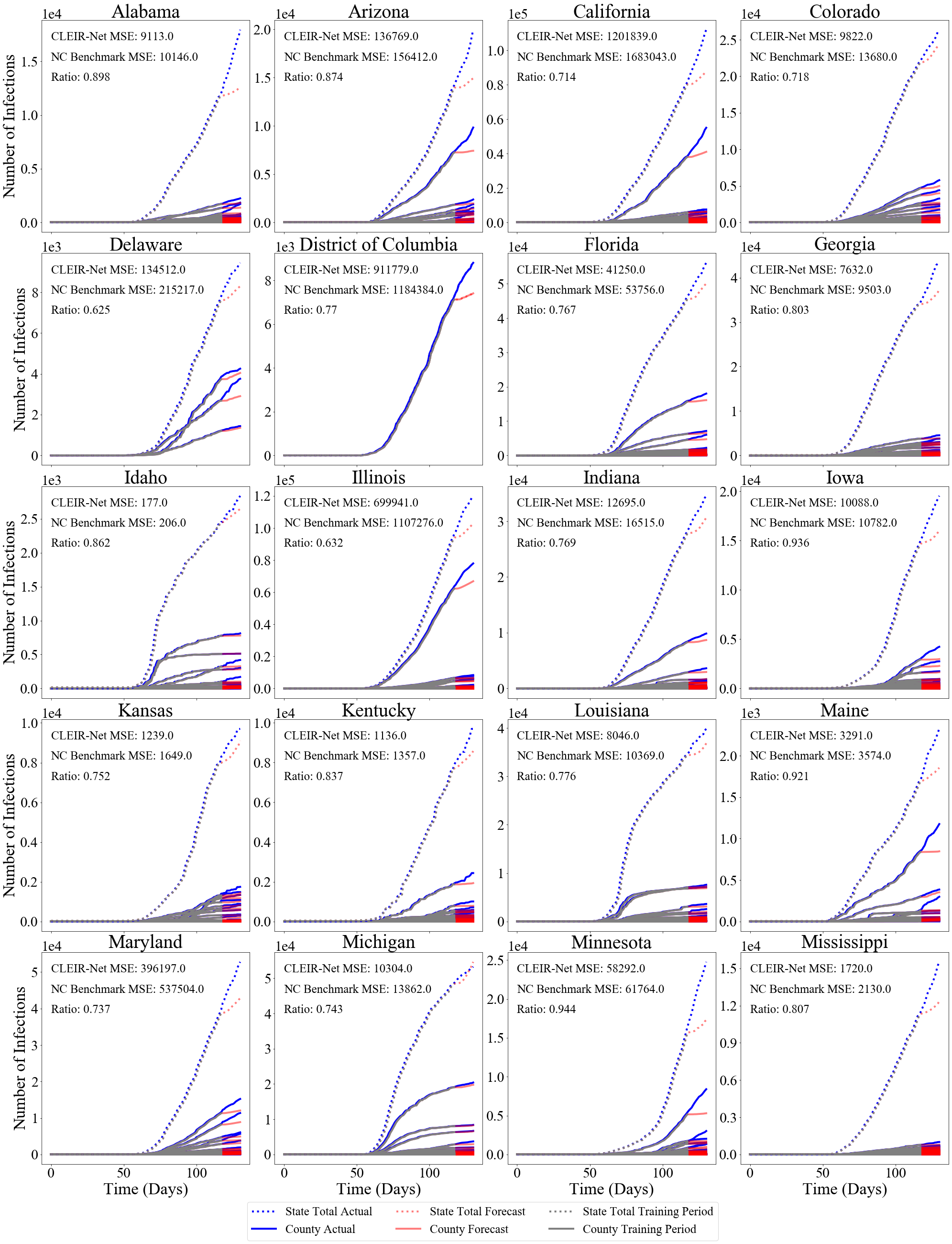}
    \label{fig:all_states:1}}
    \caption{First twenty (alphabetically) CLEIR-Net forecasts for all states and the District of Columbia, except those already shown in Figures \ref{"BestWorstMedian"} and \ref{"BestWorst5"}. See Figure \ref{fig:all_states:2} for last twenty.}
    \label{"fig:all_states"}
\end{figure}

\begin{figure}[H]
    \ContinuedFloat
    \centering
    \subfloat[]{
    \includegraphics[width=0.9\textwidth]{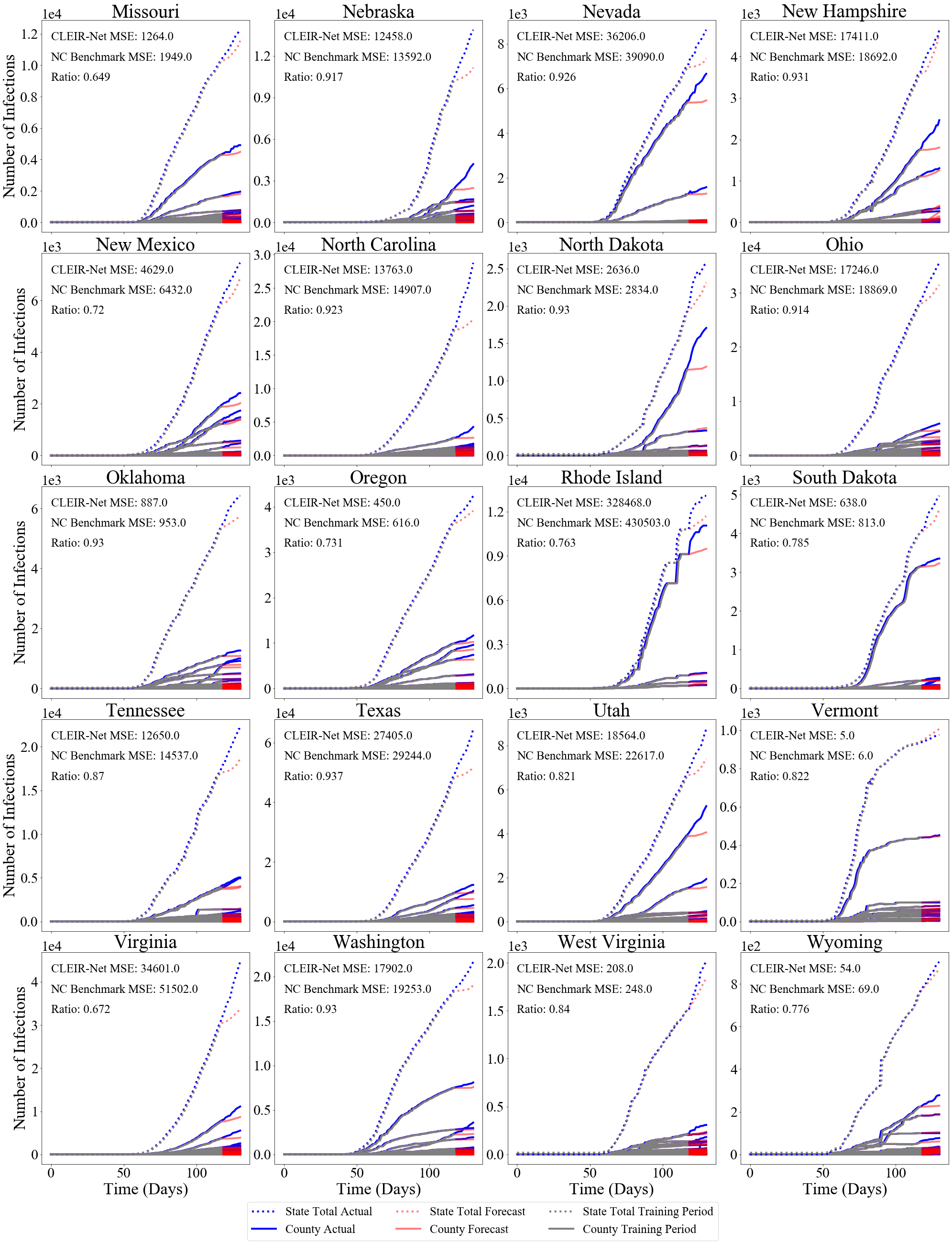}
    \label{fig:all_states:2}}
    \caption{Last twenty (alphabetically) CLEIR-Net forecasts for all states and the District of Columbia, except those already shown in Figures \ref{"BestWorstMedian"} and \ref{"BestWorst5"}. See Figure \ref{fig:all_states:1} for first twenty.}
    \label{"fig:all_states"}
\end{figure}

\end{document}